\documentclass[letterpaper, 10 pt, conference]{ieeeconf} 

\IEEEoverridecommandlockouts                              
\let\labelindent\relax
\usepackage{url}
\usepackage{graphicx}
\usepackage{svg}
\usepackage{algorithmic}
\usepackage{subfigure}
\usepackage{amsmath}
\usepackage{xcolor}
\usepackage{makecell}
\usepackage{booktabs}
\usepackage{booktabs,multirow}
\usepackage{balance}
\usepackage{listings}
\usepackage{booktabs}
\usepackage{pifont}
\usepackage{enumitem}
\usepackage[ruled,vlined,linesnumbered]{algorithm2e}
\usepackage{bm}
\usepackage{array}
\usepackage{ragged2e}
\usepackage{tabularray}
\usepackage[normalem]{ulem}
\usepackage{enumitem}
\usepackage{minted}
\usepackage{soul}
\usepackage[T1]{fontenc}
\usepackage[margin=2.5cm]{geometry}
\usepackage{tikz}

\newcommand{\blackcircled}[1]{%
  \tikz[baseline=(char.base)]{
    \node[shape=circle, fill=black, text=white,
          inner sep=0pt, minimum size=0.8em,
          font=\scriptsize\sffamily\bfseries] (char) {#1};
  }%
}

\newcommand{\ie}{\textit{i}.\textit{e}.}
\newcommand{\eg}{\textit{e}.\textit{g}.}
\newcommand{\etc}{\textit{etc}}

\definecolor{vsBg}{HTML}{FFFFFF}        
\definecolor{vsText}{HTML}{000000}      
\definecolor{vsKeyword}{HTML}{0000FF}   
\definecolor{vsControl}{HTML}{AF00DB}   
\definecolor{vsType}{HTML}{267F99}      
\definecolor{vsFunc}{HTML}{795E26}      
\definecolor{vsVar}{HTML}{001080}       
\definecolor{vsConst}{HTML}{0070C1}     
\definecolor{vsComment}{HTML}{008000}   
\definecolor{vsString}{HTML}{A31515}    
\definecolor{vsNumber}{HTML}{098658}    
\definecolor{vsLineNo}{HTML}{237893}    
\definecolor{vsRuleBg}{HTML}{F3F3F3}    
\lstdefinestyle{vscodeLightPlus}{
  language=C,
  backgroundcolor=\color{vsBg},
  basicstyle=\ttfamily\scriptsize\color{vsText},
  commentstyle=\color{vsComment},
  stringstyle=\color{vsString},
  numberstyle=\footnotesize\color{vsLineNo},
  keywordstyle=\color{vsKeyword},
  deletekeywords={if,else,for,while,do,switch,case,default,break,
                  continue,return,goto},
  morekeywords=[2]{if,else,for,while,do,switch,case,default,break,
                   continue,return,goto},
  keywordstyle=[2]\color{vsControl},
  morekeywords=[3]{rcl_ret_t,rcl_publisher_t,rmw_publisher_allocation_t,
                   rmw_ret_t,size_t,int8_t,uint8_t,int16_t,uint16_t,
                   int32_t,uint32_t,int64_t,uint64_t},
  keywordstyle=[3]\color{vsType},
  morekeywords=[4]{RCL_RET_OK,RCL_RET_ERROR,RCL_RET_PUBLISHER_INVALID,
                   RCL_RET_INVALID_ARGUMENT,RMW_RET_OK,NULL},
  keywordstyle=[4]\color{vsConst},
  identifierstyle=\color{vsVar},
  emphstyle={[5]\color{vsFunc}},
  emph={[5]rcl_publish,rcl_publisher_is_valid,rmw_publish,
        rmw_get_error_string,RCUTILS_CAN_RETURN_WITH_ERROR_OF,
        RCL_CHECK_ARGUMENT_FOR_NULL,RCL_SET_ERROR_MSG,
        TRACETOOLS_TRACEPOINT},
  numbers=left,
  numbersep=12pt,
  xleftmargin=2.2em,
  framexleftmargin=2.2em,
  frame=single,
  rulecolor=\color{vsRuleBg},
  showstringspaces=false,
  breaklines=true,
  breakatwhitespace=false,
  breakindent=1em,
  tabsize=2,
  captionpos=b,
  columns=fullflexible,
  keepspaces=true,
}

\title{Seeing is Not Believing: Breaking the Physical-to-Digital Trust Boundary in Robotics}
\author{
    Leming Shen\textsuperscript{1,2}, Shikai Geng\textsuperscript{1}, Yuanqing Zheng\textsuperscript{2} and Chris Xiaoxuan Lu\textsuperscript{1}\\
    \textsuperscript{1}University College London, \textsuperscript{2}The Hong Kong Polytechnic University
    \thanks{Work done when Leming Shen was a visiting student at UCL.}\\
    \thanks{*Chris Xiaoxuan Lu is the corresponding author}
}

\begin{document}
\maketitle
\thispagestyle{empty}
\pagestyle{empty}

\begin{abstract}
In multi-robot collaboration, task handovers rely on downstream verifiers performing remote attestation, which inspects sensor telemetry to ensure a robot's physical behavior strictly matches its assigned task. But can this telemetry be trusted? We show that it often cannot. In this paper, we uncover a severe vulnerability in Robot Operating System (ROS) 2: by modifying a single environment variable, an adversary can execute a pre-built hook to covertly intercept and inject both telemetry and control signals before they are published. Consequently, adversaries can hijack a robot to perform dangerous tasks while spoofing downstream verifiers with synthesized fake telemetry. Worse still, by exploiting the widespread reliance on third-party Docker containers and auxiliary tools, attackers can distribute compromised packages embedded with these malicious hooks to launch such attacks easily. On a physical Franka Emika robotic arm running Secure ROS 2, our attack injects fabricated telemetry in real time with only around 3 ms of jitter, preserving temporal synchronization and hardware integrity while achieving an 87\% success rate even against an AI-based detector. We have responsibly disclosed these findings to the ROS 2 development team. We prepared a demo video available at \url{https://youtu.be/ExeiGqUrnhQ}.

\end{abstract}

\section{Introduction}

As robots become increasingly autonomous and interconnected, they are moving beyond isolated, pre-programmed machines toward collaborative systems that interact with humans, peer robots, and the physical world \cite{qin2026nlipscalib, zhang2026collaborative, rekabi2025lyapunov, barcis2019robots}. Such systems are increasingly deployed in applications ranging from warehouse logistics \cite{clearpathrobotics, fetchrobotics_linkedin} to space exploration \cite{nasa_viper_2025}. In these settings, successful collaboration requires not only correct task execution, but also the ability of downstream participants to \emph{verify} what another robot has actually done.

Since continuously observing every robot is impractical and laborious, downstream verifiers (\eg, human operators, peer robots, or monitoring modules) often rely on reported sensor telemetry, such as camera frames and joint states, to assess whether physical execution conforms to the intended task \cite{ghaeini2019patt}. Such telemetry may also support remote or physics-based attestation mechanisms \cite{coker2011principles}. This creates a fundamental trust boundary between a robot's \emph{physical behavior} and its \emph{digitally reported state}. However, does trustworthy perception necessarily imply trustworthy evidence of physical execution?

\begin{figure}
    \centering
    \includegraphics[width=0.45\textwidth]{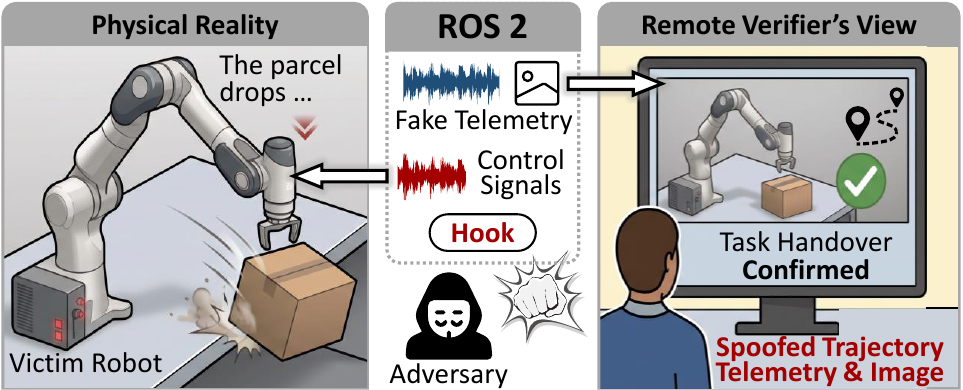}
    \caption{A compromised ROS~2 environment can decouple physical robot behavior from the telemetry observed by downstream verifiers.}
    \label{fig:blueprint}
\end{figure}

Our analysis of ROS~2 shows that the answer is no. By manipulating a single environment variable, an adversary can covertly \textbf{intercept and inject} messages before they are published, without modifying the official ROS~2\footnote{We focus on ROS~2 as it is the foundational architecture for the multibillion-dollar commercial robotics industry \cite{macenski2022robot, bonci2023robot}.} binaries (\S\ref{sec:feasibility_study}). As illustrated in Fig.~\ref{fig:blueprint}, this allows an attacker to create a "physical-digital mismatch":
its actual behavior in the physical world and a fabricated execution trace observed by downstream verifiers. In particular, adversaries can \textbf{spoof} verifiers with synthesized, high-fidelity telemetry consistent with expected behavior, manipulate control signals to facilitate \textbf{hijacking}, or inject inconsistent telemetry to launch Denial-of-Service (\textbf{DoS}) attacks. Critically, the resulting telemetry can remain syntactically valid, temporally consistent, and securely transmitted while no longer faithfully representing physical reality.

The practical risk is amplified by the dependency-heavy software ecosystem of ROS 2 \cite{sakib2025supply, xia2025investigating}. Configuring heterogeneous robotic platforms often requires developers to rely on third-party tools and pre-configured containers. Our investigation (\S\ref{sec:open_source}) of 4,429 Docker containers finds that over 4,000 are released by unverified individuals, while accounting for nearly 10.2\% of total registry activity, including up to 7.6 million pulls. An adversary can therefore distribute an apparently legitimate container that silently deploys the interception mechanism, placing malicious code within the same user-space trust domain as the robot application.

We evaluate the attack on a physical Franka Emika robotic arm \cite{haddadin2024franka} running Secure ROS~2 with native authentication enabled. Our results show that fabricated telemetry can be injected in near real time with approximately 2\,ms of transmission jitter while maintaining high-fidelity joint-state and camera images, robust temporal synchronization, and normal robot operation. These results reveal an important limitation of existing robotic security mechanisms: protecting the authenticity and integrity of telemetry during communication does not guarantee that the telemetry truthfully reflects the underlying physical execution.

We have responsibly disclosed the identified vulnerabilities, including the spoofing, hijacking, and DoS attack vectors, to the ROS~2 development team. The concerns have been acknowledged and discussed during a ROS~2 Project Management Committee (PMC) meeting.

\section{Background}
\subsection{Robot Operating System (ROS)}
ROS \cite{quigley2009ros} provides a collection of tools, libraries, and packages for robotic applications. ROS~2 extends this ecosystem toward distributed and production-oriented robotics, adopting the Data Distribution Service (DDS) standard for communication across heterogeneous platforms \cite{bonci2023robot}. It is now widely used in warehouse logistics (\eg, Clearpath \cite{clearpathrobotics}, Fetch Robotics \cite{fetchrobotics_linkedin}), autonomous agriculture, space exploration (\eg, NASA's VIPER rover \cite{nasa_viper_2025}), and autonomous driving (\eg, Autoware \cite{autoware_foundation}).

\subsection{Secure ROS 2}
\label{sec:bg_sros}
Secure ROS~2 (SROS~2) provides security mechanisms for ROS~2 applications through the underlying DDS Security standard \cite{mayoral2022sros2}. It supports certificate-based authentication, access control, and encrypted communication between ROS~2 participants. SROS~2 further organizes security credentials and policies through \emph{security enclaves}, simplifying the configuration of cryptographic keys and permissions \cite{white2016sros}. These mechanisms protect ROS~2 communication against unauthorized access, eavesdropping, and in-transit message tampering.

Importantly, these guarantees apply to messages once they enter the protected communication stack; they do not establish whether the published data faithfully represents the robot's physical state. This distinction motivates the attack studied in this work.

\subsection{ROS 2 Ecosystem and Supply-Chain Exposure}
\label{sec:open_source}
Deploying ROS~2 across heterogeneous hardware often requires third-party packages, tools, and pre-configured containers \cite{abukhalil2016deployment}, introducing potential software supply-chain exposure \cite{xia2025investigating}. To characterize this ecosystem, we crawl 4,429 ROS-related Docker images and 5,322 GitHub repositories and collect their usage statistics (\eg, the number of pulls and forks).

\begin{itemize}[leftmargin=9pt]
    \item \textbf{Long-tail Docker exposure.}
    Major verified namespaces (\eg, \texttt{ros}, \texttt{osrf}, and \texttt{moveit}) account for 89.8\% of 74.7 million observed pulls. Nevertheless, approximately 4,400 images outside these namespaces collectively account for 7.6 million pulls (10.2\%), indicating substantial use of third-party images.

    \item \textbf{A fragmented repository ecosystem.}
    The 5,322 repositories span 3,636 owners, with only 76 marked as official. More than 2,000 repositories have fewer than five stars, illustrating a large long tail of independently maintained, low-visibility projects.
\end{itemize}
These findings do not imply that third-party artifacts are malicious. Rather, they indicate that ROS~2 deployments rely on a large yet decentralized software ecosystem, providing a plausible distribution channel for the compromised artifacts considered in our threat model.

\section{Threat Model}
We consider the trust relationship between a robot's physical execution and the telemetry observed by a downstream verifier. The verifier (\eg, a human operator, monitoring component, or peer robot) is uncompromised and relies on ROS~2 telemetry (\eg, camera frames and joint states) to assess whether the robot has executed its assigned task correctly. We assume that the verifier does not have an independent trusted sensor that continuously observes the robot's physical state. As motivated in \S\ref{sec:open_source}, the attacker obtains a user-space foothold through a compromised third-party ROS~2 artifact, such as a trojanized container or package. The artifact is executed by the victim with ordinary application privileges. We do \textit{not} assume root access, kernel exploitation, modification of official ROS~2 binaries, compromise of SROS~2 credentials, or physical access to the robot.

\subsection{Attack Goals}

The attacker's objective is to break the correspondence between the robot's physical behavior and the execution evidence observed by downstream verifiers. We consider two attack goals:

\begin{itemize}[leftmargin=9pt]
    \item \textbf{Hijacking \& Spoofing.}
    The attacker alters control signals to cause the victim robot to deviate from its intended task, while replacing its outbound telemetry with synthesized data consistent with the expected execution. Consequently, the verifier may accept an incorrect physical execution as legitimate.

    \item \textbf{Denial-of-Service (DoS).}
    The attacker injects inconsistent or abnormal telemetry to disrupt the verification process, causing verifiers to reject executions, initiate repeated diagnostics, or halt the workflow.
\end{itemize}

\subsection{Attacker's Capabilities}

\noindent\textbf{System Knowledge.}
The attacker knows the ROS~2 topics and message types used by the target application (\eg. \texttt{Image}, \texttt{JointState}, \etc) and can construct structurally valid replacement messages.

\noindent\textbf{User-Space Execution.}
The attacker can execute code within the same user-space environment as the victim ROS~2 application and intercept selected telemetry and control messages before publication. The attacker cannot modify the OS kernel, official ROS~2 middleware binaries, robot firmware, or the downstream verifier.

\noindent\textbf{No SROS~2 Compromise.}
The attacker does not break DDS Security, steal cryptographic credentials, or modify messages after they enter the protected communication channel. Instead, manipulation occurs before the SROS~2 protections described in \S\ref{sec:bg_sros} take effect. Therefore, the resulting falsified messages can be authenticated and securely delivered to the verifier as legitimate traffic.

\section{Attack Surface and Exploit Mechanism}
\label{sec:feasibility_study}

This section analyzes the ROS~2 publication path and identifies a pre-publication interception point that enables our attack. We then show how user-space function interposition can manipulate messages at this boundary without modifying official ROS~2 binaries, and explain why such manipulation is not prevented by SROS~2.

\subsection{ROS~2 Publication Path}

To identify where outbound ROS~2 messages can be intercepted before transmission, we trace the publication path of a minimal Python-based ROS~2 publisher. Using step-through debugging, we trace execution from \texttt{rclpy} into the ROS~2 client library \texttt{librcl.so}. We further use radare2 \cite{radare2book} to inspect the binary implementation and confirm that publication ultimately passes through the core \texttt{rcl} library \cite{rcl}, which contains a \texttt{rcl\_publish} function as shown below:

\noindent
\begin{minipage}{0.48\textwidth}
\vspace{4pt}
\begin{lstlisting}[style=vscodeLightPlus, language=C]
rcl_ret_t rcl_publish(
  const rcl_publisher_t * publisher,
  const void * ros_message,
  rmw_publisher_allocation_t * allocation)
{
  if (!rcl_publisher_is_valid(publisher))
    return RCL_RET_PUBLISHER_INVALID;

  RCL_CHECK_ARGUMENT_FOR_NULL(
      ros_message, RCL_RET_INVALID_ARGUMENT);

  if (rmw_publish(
        publisher->impl->rmw_handle,
        ros_message,
        allocation) != RMW_RET_OK) {
    return RCL_RET_ERROR;
  }

  return RCL_RET_OK;
}
\end{lstlisting}
\end{minipage}

\texttt{rcl\_publish} receives a publisher handle and the corresponding message, and subsequently delegates transmission to \texttt{rmw\_publish}, which interfaces with the underlying ROS~2 middleware. Therefore, \texttt{rcl\_publish} forms a common pre-middleware publication boundary: message contents remain directly accessible before being handed to the communication layer.

\begin{figure}
    \centering
    \includegraphics[width=0.48\textwidth]{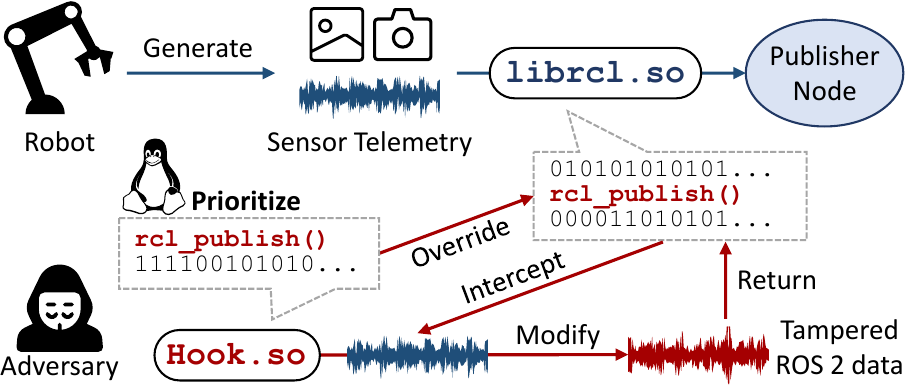}
    \caption{Pre-publication interception of ROS~2 messages through
    user-space function interposition.}
    \label{fig:attack_rationale}
\end{figure}

\subsection{Pre-Publication Function Interposition}

A naive approach to manipulating messages at this boundary would be to modify and recompile \texttt{librcl.so}. However, doing so alters the official ROS~2 binary and can be exposed by software-integrity checks. We therefore ask: \textit{can the publication path be intercepted without modifying ROS~2 itself?}

Linux provides dynamic function interposition through the \texttt{LD\_PRELOAD} environment variable, which causes a specified shared library to be loaded before ordinary dynamically linked libraries. If the preloaded library exports the same symbol as a subsequently loaded library, calls to that symbol can instead be resolved to the preloaded implementation.

We exploit this mechanism at the ROS~2 publication boundary. Specifically, an attacker supplies a malicious shared library, denoted \texttt{Hook.so}, that exports its own implementation of \texttt{rcl\_publish}. When the affected ROS~2 process starts, calls to \texttt{rcl\_publish} are first redirected to the hook, allowing selected messages to be inspected or modified before they are forwarded to the legitimate implementation in \texttt{librcl.so}. Importantly, the official ROS~2 binaries remain unchanged.

As illustrated in Fig.~\ref{fig:attack_rationale}, the malicious hook first intercepts the call to \texttt{rcl\_publish}, identifies whether the publisher node corresponds to a target topic (\eg, \texttt{/franka/joint\_states}), and modifies the message if necessary. It then invokes the legitimate \texttt{rcl\_publish}, allowing the altered payload to continue through the normal ROS~2 publication path. The same interception primitive applies to both telemetry and control publishers, enabling the spoofing and hijacking components described in the following section. This essential interception logic can be summarized as follows:

\begin{figure*} 
    \centering
    \includegraphics[width=0.9\textwidth]{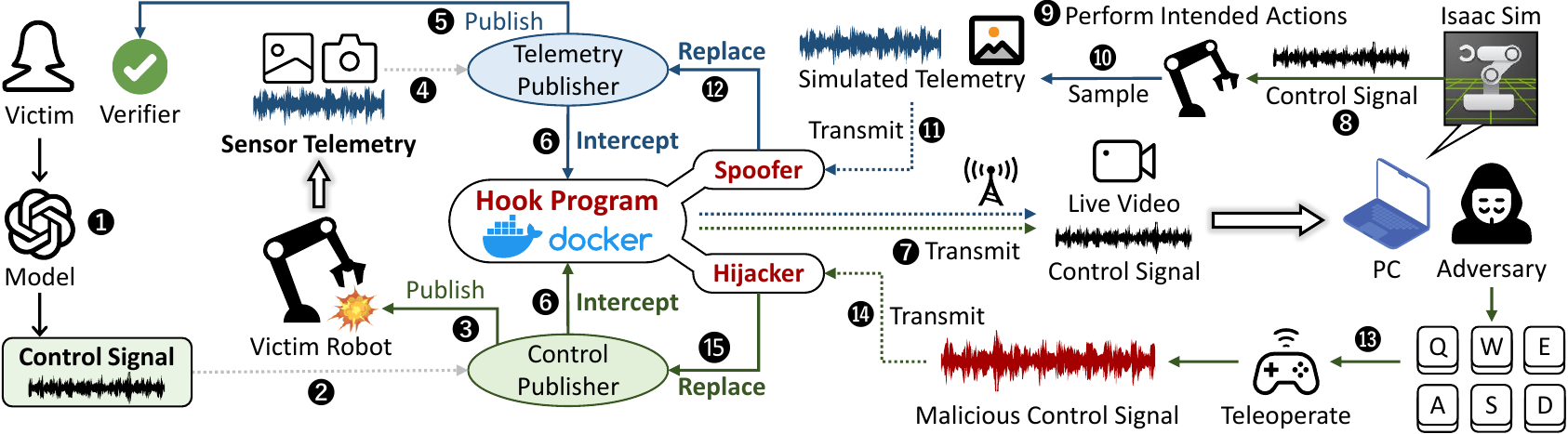}
    \caption{The attack pipeline. The malicious hook executes with only user-level privileges and intercepts messages before they enter the protected ROS~2 communication stack.} 
    \label{fig:attack_pipeline} 
\end{figure*}

\noindent
\begin{minipage}{0.48\textwidth}
\vspace{4pt}
\begin{lstlisting}[style=vscodeLightPlus, language=C]
extern "C" int rcl_publish(...) {
  original = dlsym(RTLD_NEXT, "rcl_publish");

  if (is_target_topic(publisher)) {
    modify(ros_message);
  }

  return original(
      publisher, ros_message, allocation);
}
\end{lstlisting}
\end{minipage}

\subsection{Why SROS~2 Does Not Prevent the Attack}

SROS~2 protects ROS~2 communication through authentication, access control, and encryption at the DDS layer. However, our interception point lies \emph{upstream} of these protections: the hook modifies the message before it is passed from \texttt{rcl\_publish} to the underlying RMW/DDS communication stack.

Consequently, the attacker does not need to break SROS~2, recover cryptographic credentials, or tamper with protected network traffic. Instead, the falsified message is further authenticated, encrypted, and transmitted through the normal SROS~2 pipeline as legitimate ROS~2 traffic. Thus, the attack does not violate SROS~2's cryptographic guarantees; rather, it exploits a pre-publication trust boundary that those guarantees do not cover.

\section{Detailed Design}

\subsection{Attack Pipeline}
Fig.~\ref{fig:attack_pipeline} illustrates our attack pipeline.
Normally, when a user issues a high-level command (\eg, "pick and place"), a policy model first translates it into robotic control signals (\blackcircled{1}) and then creates a publisher node (\blackcircled{2}) to publish these signals to a robot (\blackcircled{3}). In the meantime, the captured robot telemetry (\blackcircled{4}) is published by another publisher node to a downstream verifier for monitoring and action verification (\blackcircled{5}).

The malicious hook consists of two main modules: a \textit{hijacker} to control the victim robot and a \textit{spoofer} to deceive the downstream \textbf{verifier}. They both operate based on the novel attack channel we discovered via the \texttt{LD\_PRELOAD} environment variable. Specifically, the hook first intercepts the outbound data (\blackcircled{6}), including both sensor telemetry and control signals, and transmits them to the adversary's PC (\blackcircled{7}). The PC runs a high-fidelity robot simulator, Isaac Sim \cite{gao2026nvidia}, which immediately applies the intercepted control signals to a virtual robot (\blackcircled{8}) to perform the intended task (\blackcircled{9}). Simultaneously, the PC captures the simulated robot's telemetry (\blackcircled{10}) that perfectly matches the legitimate telemetry's type and structure. The synthesized telemetry is then sent back to the spoofer (\blackcircled{11}) that seamlessly replaces the real telemetry (\blackcircled{12}) before publication, thereby spoofing the downstream verifier with synthesized, seemingly benign telemetry. Concurrently, to hijack the robot, the attacker monitors its real-time state via received camera frames (\blackcircled{7}) and teleoperates it via keyboard inputs (\blackcircled{13}). These inputs are translated into malicious control signals, forwarded to the hijacker (\blackcircled{14}), and injected in place of the original control signals before publication. (\blackcircled{15}).

Because the falsified telemetry is also consumed by subsequent policy iterations, the policy continues planning against the simulated state rather than the true physical state. Consequently, the physical robot and its digital twin evolve as \textit{two parallel execution traces}: the attacker controls the former, while the policy and verifier remain coupled to the latter.

\subsection{Generating Realistic Spoofed Telemetry}
\label{sec:synthesize}
While generating alternative control signals to hijack the physical robot is relatively simple, generating \emph{realistic spoofed telemetry} that can consistently fool a downstream verifier is considerably more challenging. A high-fidelity simulator such as Isaac Sim can reproduce the intended task trajectory, but its raw telemetry still differs from that of a physical robot in subtle yet detectable ways. In particular, during stationary phases, a real robot continues to exhibit small fluctuations caused by controller dynamics, mechanical vibration, and friction, whereas simulated joint states remain nearly constant.

A straightforward solution is to add independent Gaussian noise to the simulated telemetry. However, this fails to capture two key properties of real robot signals. First, real telemetry is temporally structured: its frequency spectrum contains low-frequency drift and characteristic vibration components rather than spectrally flat noise. Second, different telemetry channels are physically coupled. Joint positions, velocities, and torques arise from the same underlying robot motion and therefore cannot be perturbed independently without introducing inconsistencies that a verifier could detect.

Our key idea is therefore to perturb the \emph{underlying simulated motion} rather than individual telemetry channels. Let $q_j^\star(t)$ denote the trajectory produced by the simulator for joint $j$, where $j\in\{1,\ldots,N\}$ and $N$ is the number of robot joints. We generate a small latent perturbation $\delta q_j(t)$ and construct the spoofed physical-like trajectory as
\begin{equation}
q_j(t)=q_j^\star(t)+\delta q_j(t).
\end{equation}
We refer to $\delta q_j(t)$ as the \emph{latent micro-motion}. As all telemetry channels are subsequently derived from this shared trajectory, they remain mutually consistent and produce hardware-like temporal and cross-joint structure.

\noindent\textbf{Latent micro-motion.}
We model $\delta q_j(t)$ as a combination of vibration, slow drift, and
occasional stick--slip effects:
\begin{equation}
\delta q_j(t)=
\underbrace{\sum_{m=1}^{M}s_{j,m}(t)}_{\text{vibration}}
+
\underbrace{d_j(t)}_{\text{drift}}
+
\underbrace{J_j(t)}_{\text{stick-slip}},
\end{equation}
where $M$ is the number of vibration modes and $s_{j,m}(t)$ denotes the $m$-th vibration mode of joint $j$. Each mode follows a damped stochastic oscillator:
\begin{equation}
\ddot{s}_{j,m}
+
2\zeta_{j,m}\omega_{j,m}\dot{s}_{j,m}
+
\omega_{j,m}^{2}s_{j,m}
=
b_{j,m}w_{j,m}(t),
\end{equation}
where $\zeta_{j,m}$ is the damping ratio, $\omega_{j,m}$ is the natural frequency, $b_{j,m}$ controls the disturbance magnitude, and $w_{j,m}(t)$ is a zero-mean stochastic excitation. The oscillator terms model structural vibration and controller-induced oscillation. The term $d_j(t)$ is a slowly varying Ornstein--Uhlenbeck process \cite{uhlenbeck1930theory} that captures low-frequency drift, while $J_j(t)$ models occasional small discontinuities associated with stick--slip friction.

\noindent\textbf{Inter-joint coupling.}
To preserve realistic inter-joint correlation, the perturbations are generated jointly rather than independently for each joint. Let $q^\star(t)=[q_1^\star(t),\ldots,q_N^\star(t)]^\top$ denote the simulated joint configuration and let $\mathbf{M}(q^\star)$ denote the corresponding robot mass matrix. We shape the covariance of the stochastic joint disturbances as
\begin{equation}
\Sigma_w
=
\mathbf{M}(q^\star)^{-1}
\Sigma_u
\mathbf{M}(q^\star)^{-\top},
\end{equation}
where $\Sigma_u$ is the covariance of an underlying shared disturbance and $\Sigma_w$ is the resulting joint-space disturbance covariance. This produces coordinated micro-motion whose coupling depends on arm configurations.

\begin{figure}
    \centering
    \includegraphics[width=0.45\textwidth]{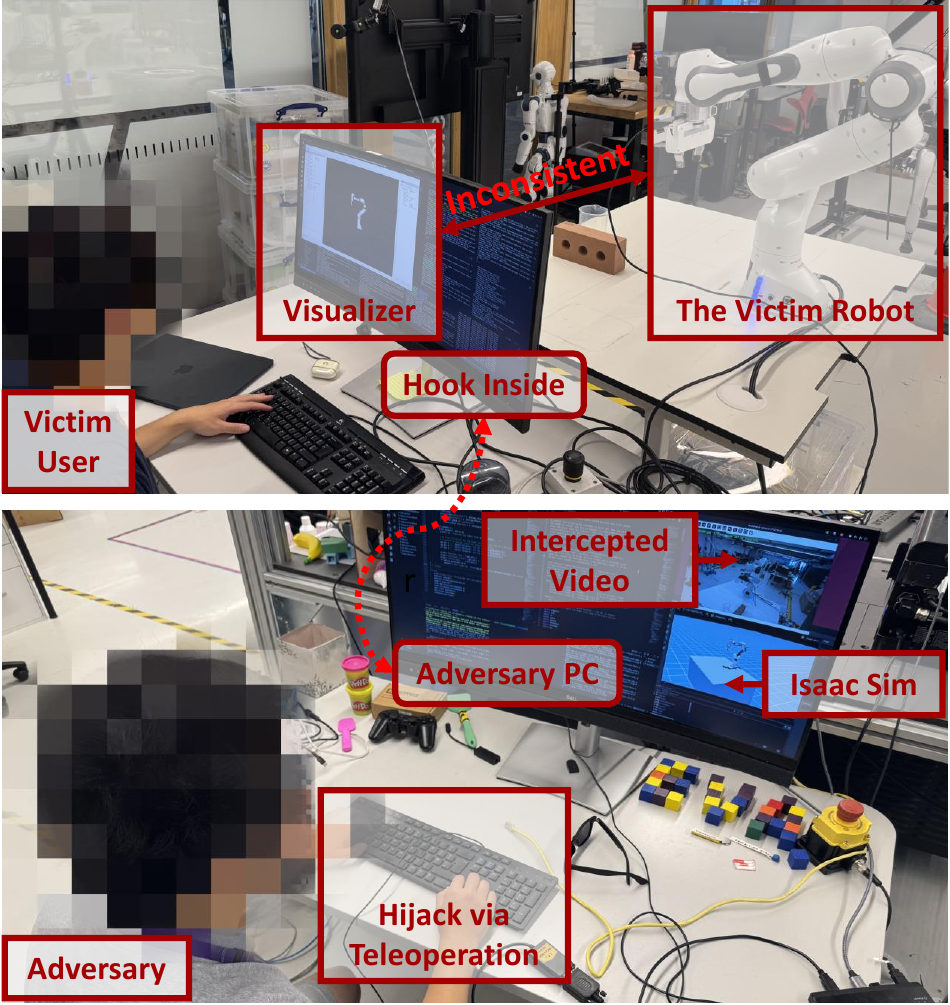}
    \caption{Experiment setup with a Franka arm.}
    \label{fig:experiment_setup}
\end{figure}

\noindent\textbf{Cross-channel physical consistency.}
Let
$q(t)=q^\star(t)+\delta q(t)$ denote the perturbed joint trajectory, where $\delta q(t)=[\delta q_1(t),\ldots,\delta q_N(t)]^\top$. Position and velocity telemetry are derived from this same trajectory, while torque is computed using the robot's inverse dynamics:
\begin{equation}
\boldsymbol{\tau}^{\mathrm{model}}
=
\mathbf{M}(q)\ddot q
+
\mathbf{C}(q,\dot q)\dot q
+
\mathbf{g}(q)
+
\mathbf{f}_{\mathrm{fric}}(\dot q),
\end{equation}
where $\mathbf{C}(q,\dot q)\dot q$ denotes Coriolis and centrifugal effects \cite{lewowski1999measurement}, $\mathbf{g}(q)$ denotes gravity compensation, and $\mathbf{f}_{\mathrm{fric}}(\dot q)$ models near-standstill friction \cite{armstrong1994survey}. Because position, velocity, and torque are all derived from the same perturbed trajectory, the resulting spoofed telemetry preserves temporal structure, inter-joint correlation, and physical consistency across channels.

In practice, the parameters governing vibration, drift, disturbance coupling, and friction
can be calibrated from a short recording of quiescent physical-robot telemetry. This could be achieved during the robot's homing process, providing an opportunity for our hook to intercept a short window of static telemetry for calibration.

\section{Evaluations}

\begin{figure*}
    \centering
    \subfigure[Delay \& jitter]{
        \includegraphics[width=0.23\textwidth]{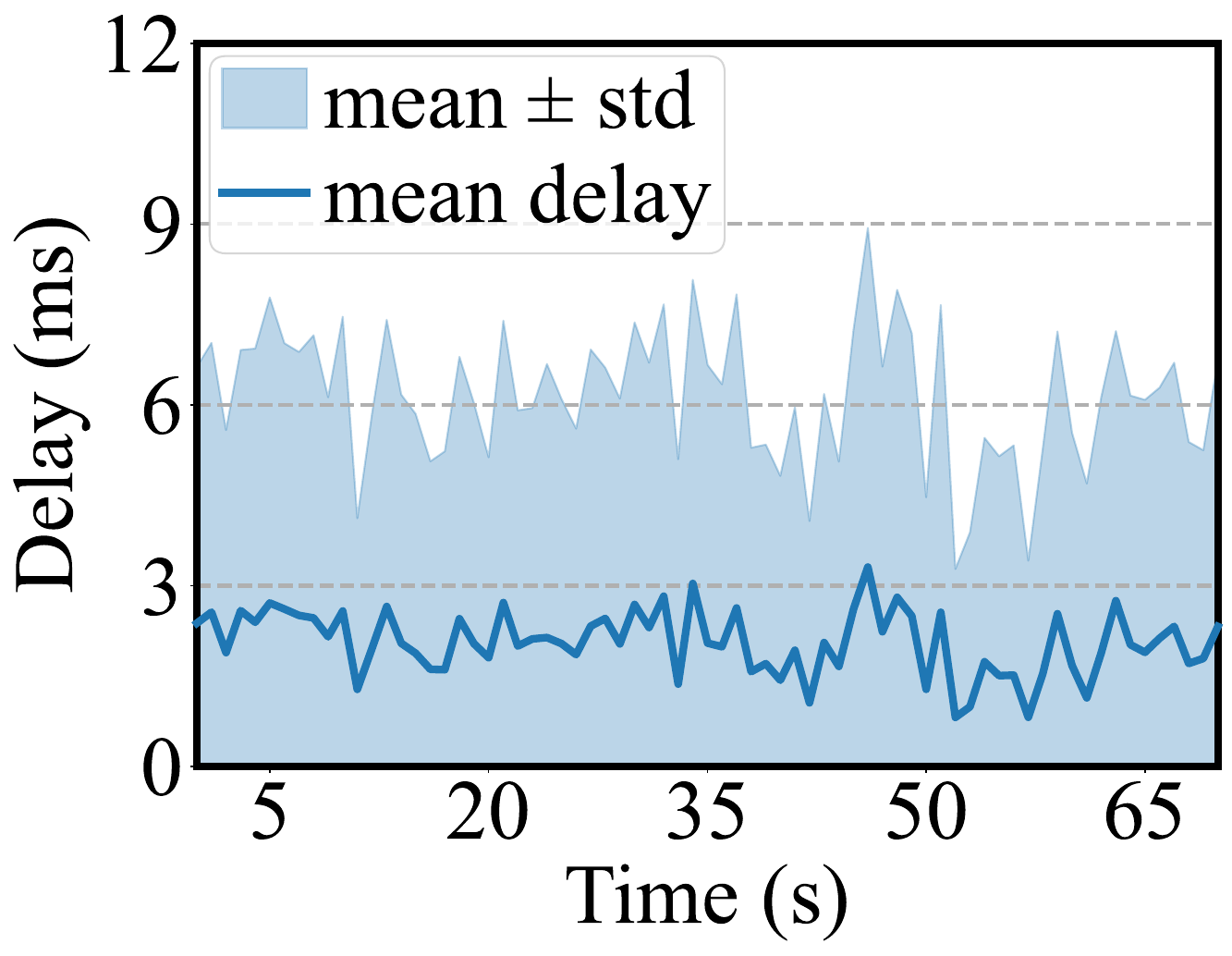}
        \label{fig:latency_performance}
    }
    \subfigure[RMSE of positions]{
        \includegraphics[width=0.23\textwidth]{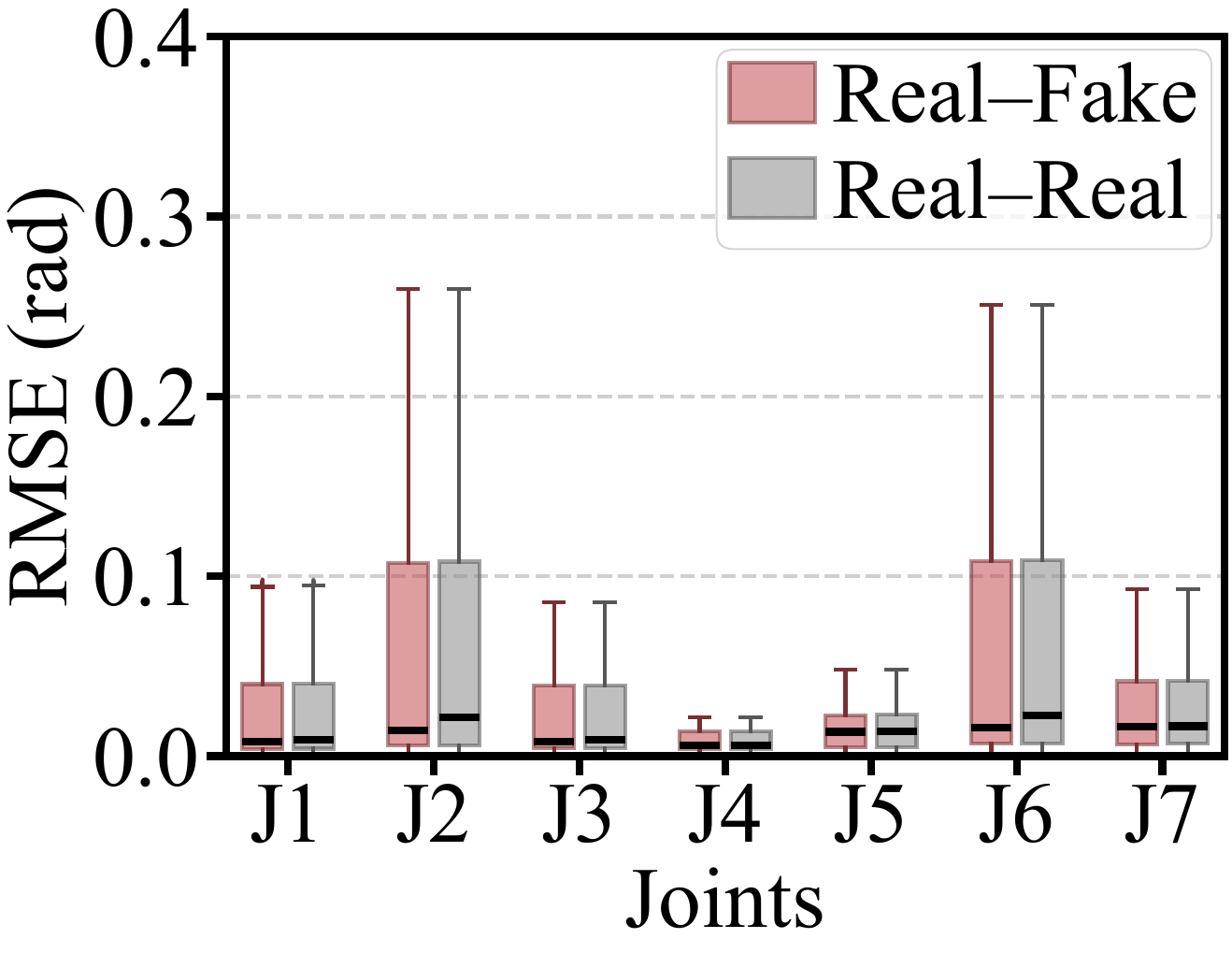}
        \label{fig:position_rmse}
    }
    \centering
    \subfigure[DTW of velocities]{
        \includegraphics[width=0.23\textwidth]{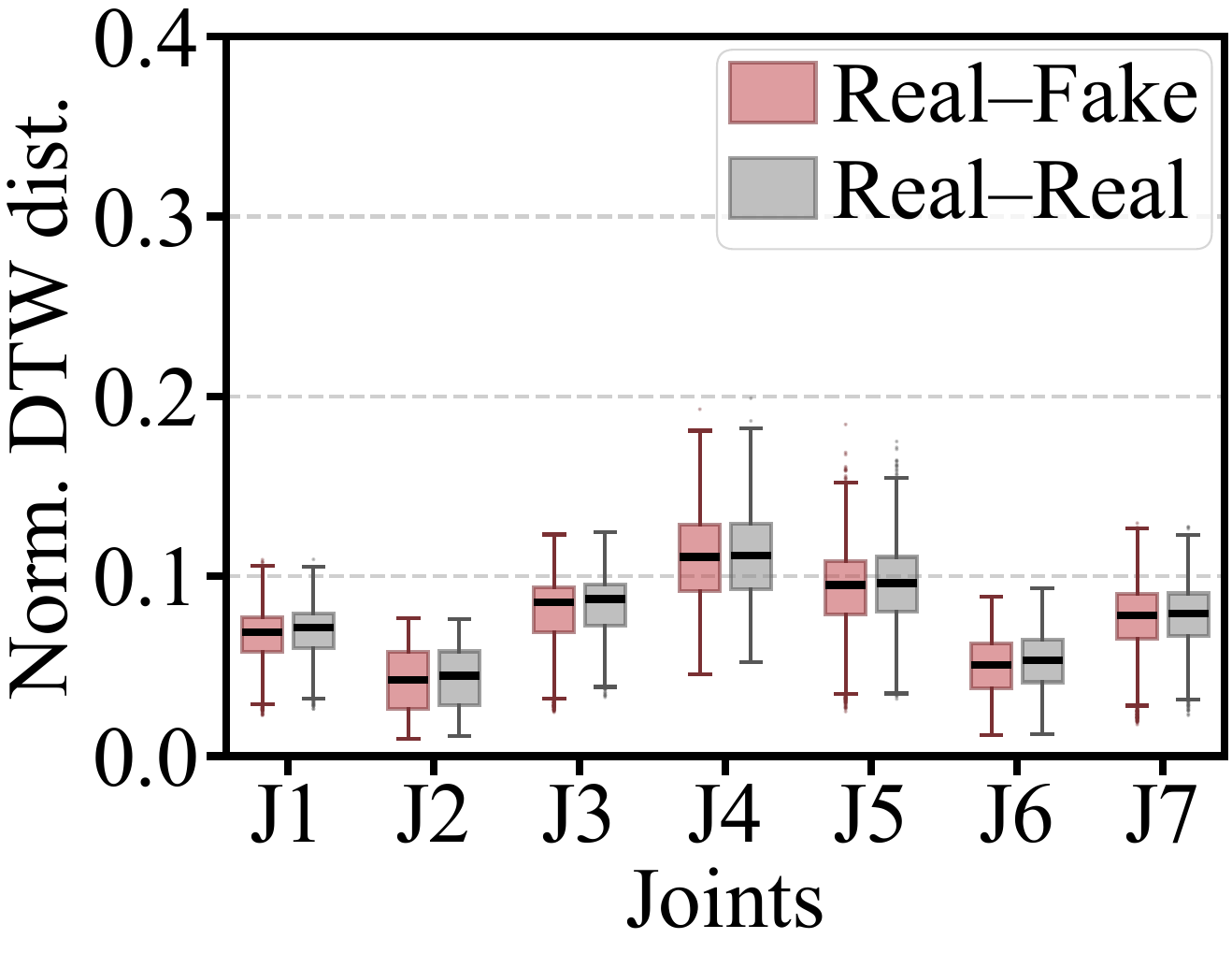}
        \label{fig:velocity_dtw}
    }
    \subfigure[Std of torques]{
        \includegraphics[width=0.23\textwidth]{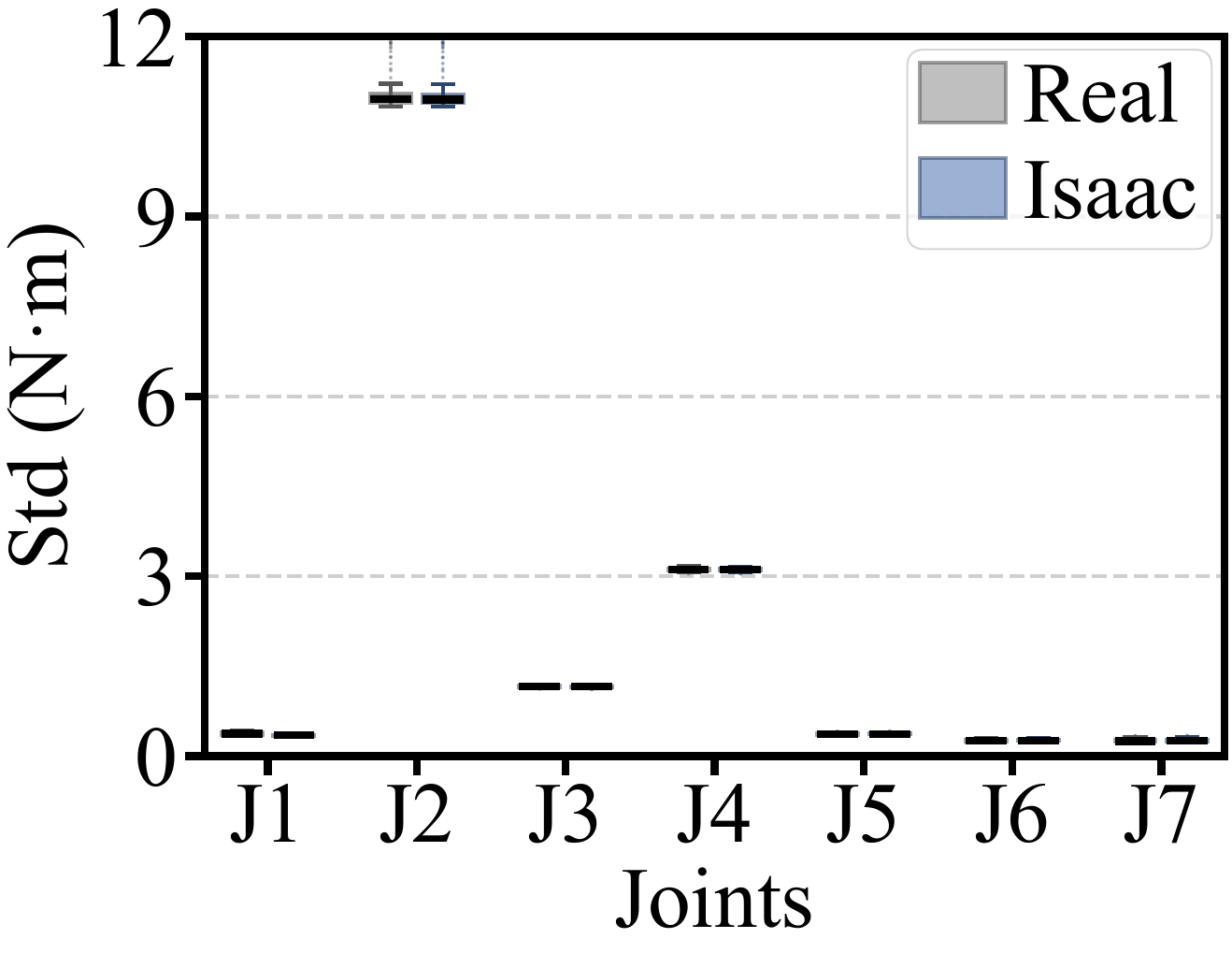}
        \label{fig:effort_std}
    }
    \caption{The real-world performance of our attack.}
    \label{fig:pretrained_model_evaluation}
\end{figure*}

\subsection{Experiment Setup}
We conduct experiments running ROS~2 Humble \cite{macenski2022robot} (one widely used version) on an Ubuntu server (\textit{victim}), equipped with an AMD Ryzen Threadripper Pro CPU and an NVIDIA RTX A6000 GPU. The victim is linked to a Franka Emika robot arm \cite{haddadin2024franka} and runs an RViz visualizer to monitor real-time joint states. The \textit{attacker} is running Isaac Sim on another Ubuntu server with the same configurations. Note that throughout the attack, no root user privileges are required, and the ROS~2 package stays intact. The detailed setup is shown in Fig.~\ref{fig:experiment_setup}.

We aim to answer the following research questions:
\begin{itemize}[leftmargin=9pt]
    \item \textbf{RQ1}: Can the attack hijack the robot while evading downstream verification under ROS~2 and SROS~2?
    \item \textbf{RQ2}: What's the real-time performance of the attack?
    \item \textbf{RQ3}: What is the fidelity of the synthesized telemetry?
    \item \textbf{RQ4}: Can the attack extend to multimodal telemetry?
    \item \textbf{RQ5}: How does each technical component contribute to the fidelity of the synthesized telemetry?
\end{itemize}

\subsection{Overall Attack Success Rate (RQ1)}
\label{sec:overall}
We repeat the "pick and place" task 100 times with and without SROS~2, respectively, and report the average attack success rate. An attack is considered successful if the robot is hijacked to execute unauthorized actions, while the victim's RViz visualizer continues to display normal task execution.
Results show that in both cases, the attack achieves a \textbf{100\%} success rate, demonstrating that our discovered attack channel reliably intercepts and manipulates ROS~2 messages before SROS~2 acts. 

To further test if the forged telemetry defeats representative downstream verification methods, we implement: \textbf{D1}: A velocity range check that learns the position, velocity, and effort bounds from real recordings and flags any sample exceeding these limits, including implausibly fast motion. \textbf{D2}: A position–velocity consistency check that verifies if the velocity mathematically aligns with the temporal change in position over time, thereby detecting variables perturbed independently rather than derived from continuous motion. \textbf{D3}: A cross-joint correlation analysis that checks if joint jitter exhibits the characteristic correlation patterns produced by a shared physical structure, which independent per-joint noise cannot mimic. \textbf{D4}: a temporal anomaly detector that trains a small autoregressive model on the genuine telemetry to flag those whose temporal structure markedly deviates from the predictability of real hardware.

We replay our attack 100 times against each detector and report the bypass rate: \textbf{98\%}, \textbf{100\%}, \textbf{95\%}, and \textbf{87\%}. The perfect evasion of D2 follows directly from our design: as every channel is derived from a single perturbed trajectory (\S\ref{sec:synthesize}), the reported velocity is exactly the time-derivative of the reported position, leaving no kinematic inconsistency to expose. The high bypass rates of D1 and D3 likewise reflect that the synthesized telemetry stays within learned physical limits, and that our mass-matrix coupling reproduces the correlation structure of a shared kinematic chain that independent per-joint noise cannot. D4 is the strongest defense, flagging 13\% of attacks, as the fine-grained temporal predictability of real hardware is highly difficult to reproduce exactly. Even so, an 87\% bypass rate leaves the large majority of attacks undetected. Taken together, no single detector reliably separates fabricated from genuine telemetry, indicating that statistical verification alone cannot restore trust once the pre-publication boundary is breached.

\subsection{Real-Time Performance (RQ2)}
Since the synthesized telemetry is sent to the victim over the network, we consider three evaluation metrics: 1) \textit{Delay}, defined as the time elapsed between the invocation of \texttt{rcl\_publish} and the delivery of the payload to the DDS middleware layer; 2) \textit{Jitter}, measured as the standard deviation of the latency to quantify transmission variance; and 3) \textit{Message Drop Rate}, representing the percentage of dropped telemetry samples.

\begin{figure*}
    \centering
    \subfigure[The initial state (Isaac Sim)]{
        \includegraphics[width=0.225\textwidth]{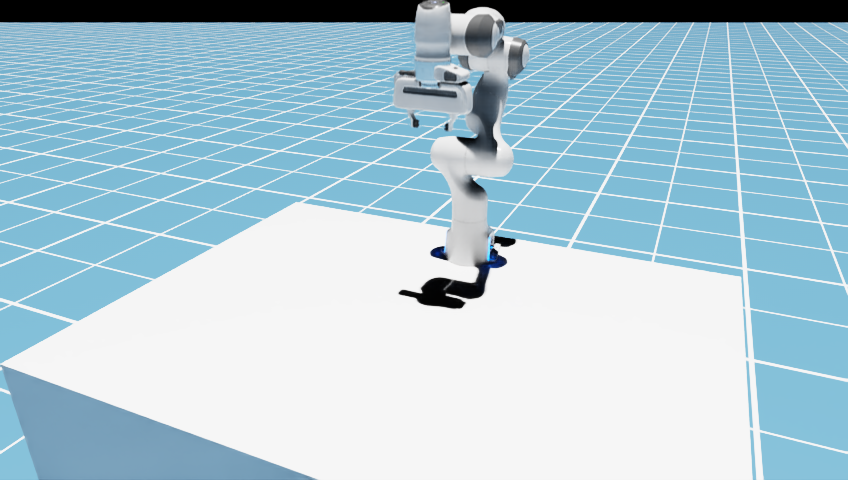}
        \label{fig:isaac_initial}
    }
    \centering
    \subfigure[The initial state (real)]{
        \includegraphics[width=0.225\textwidth]{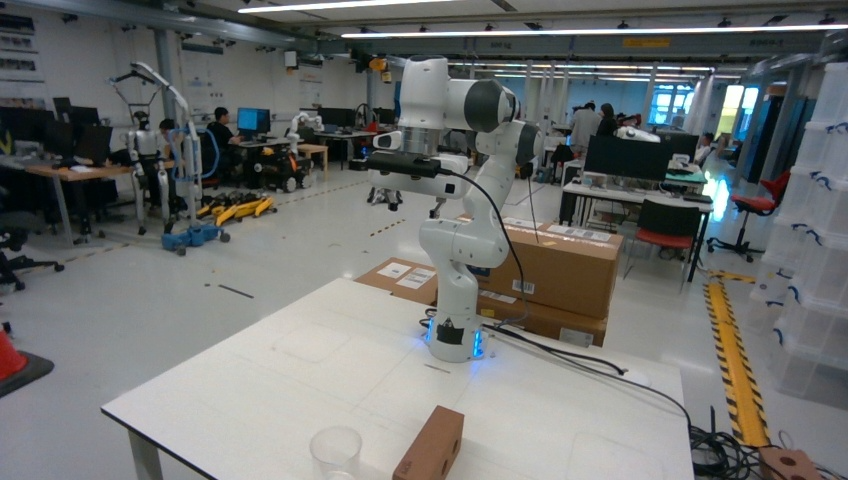}
        \label{fig:initial_state}
    }
    \centering
    \subfigure[The final state (Isaac Sim)]{
        \includegraphics[width=0.225\textwidth]{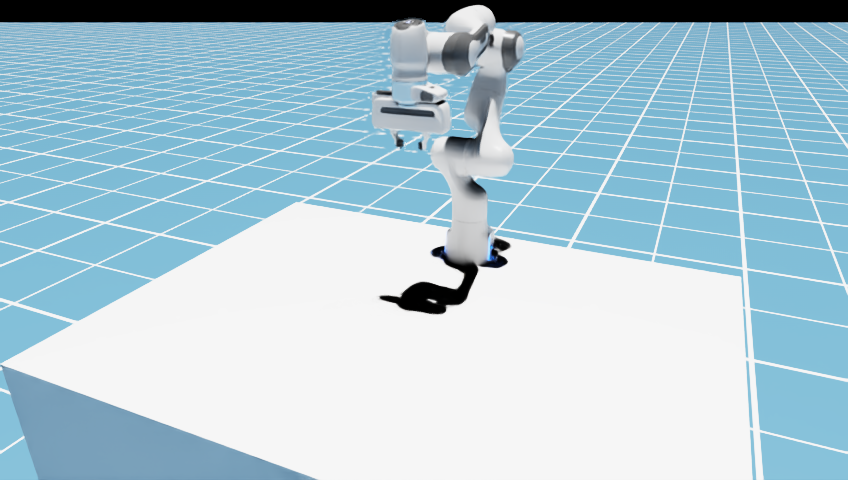}
        \label{fig:isaac_final}
    }
    \centering
    \subfigure[The final state (fake)]{
        \includegraphics[width=0.225\textwidth]{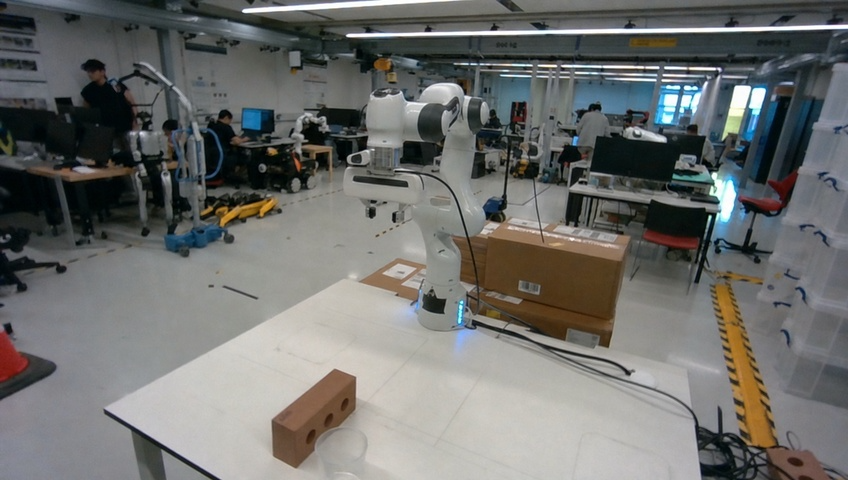}
        \label{fig:synthesized_state}
    }
    \centering
    \subfigure[The initial state (Isaac Sim)]{
        \includegraphics[width=0.225\textwidth]{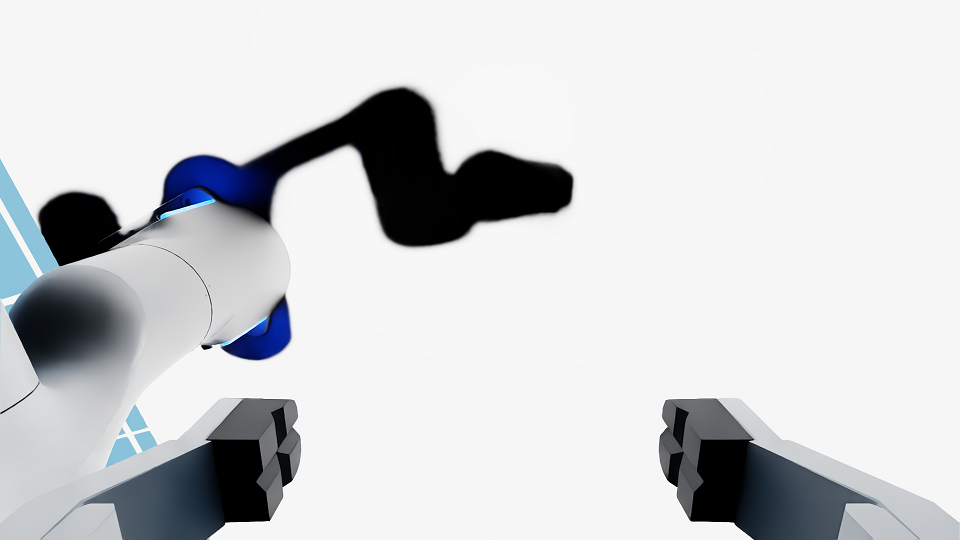}
        \label{fig:isaac_initial_zed}
    }
    \centering
    \subfigure[The initial state (real)]{
        \includegraphics[width=0.225\textwidth]{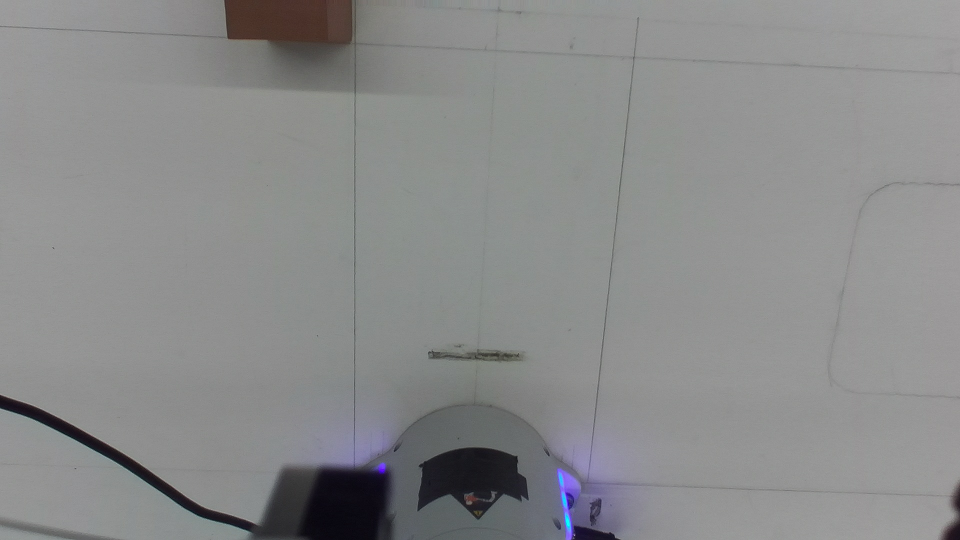}
        \label{fig:initial_state_zed}
    }
    \centering
    \subfigure[The final state (Isaac Sim)]{
        \includegraphics[width=0.225\textwidth]{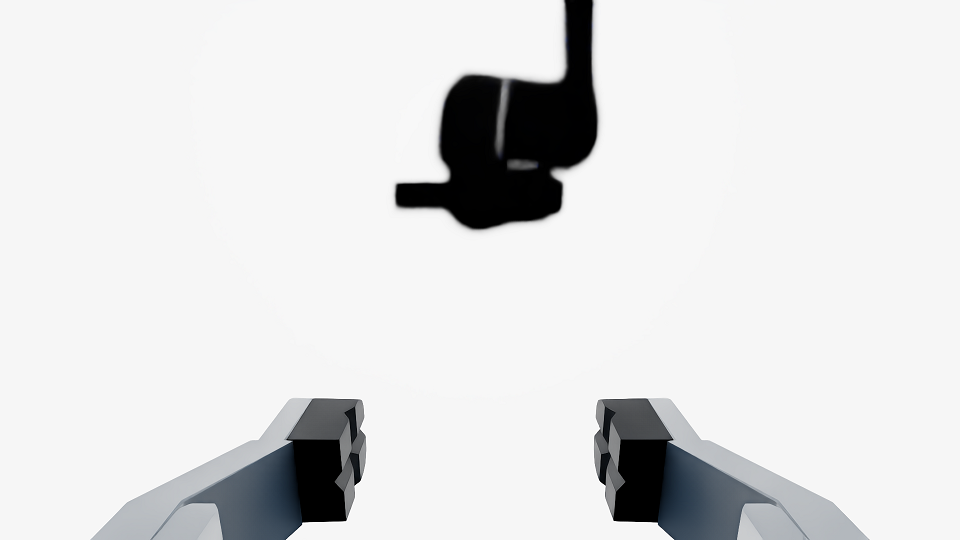}
        \label{fig:isaac_final_zed}
    }
    \centering
    \subfigure[The final state (fake)]{
        \includegraphics[width=0.225\textwidth]{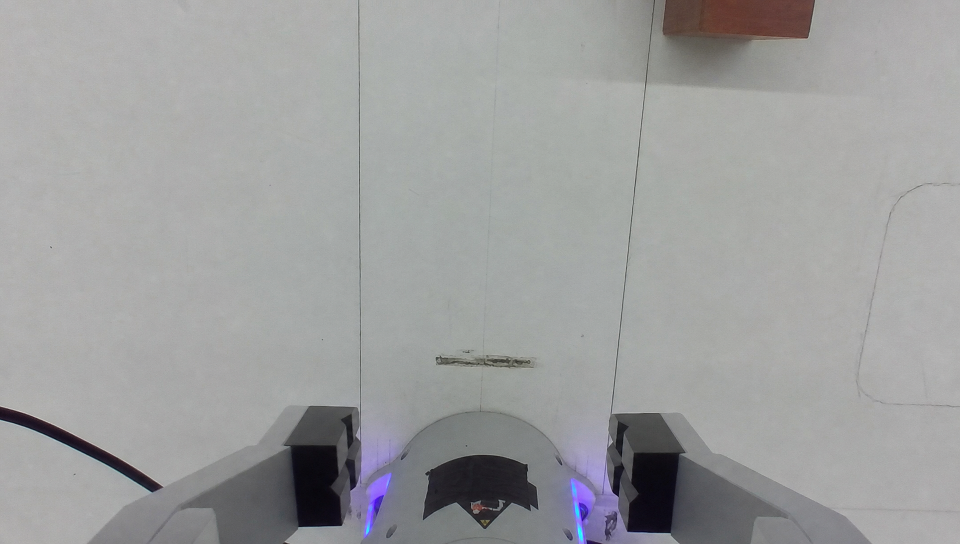}
        \label{fig:synthesized_state_zed}
    }
    \caption{The simulated images (a, c, e, g), the real image (b, f), and the synthesized image (d, h). The first row corresponds to a third-person perspective camera (Intel RealSense \cite{keselman2017intel}) and the second row corresponds to a wrist camera (ZED Mini \cite{stereolabs_zed_mini}).}
    \label{fig:multimodal}
\end{figure*}

We report the above metrics over a 1-minute interval, during which the robot holds still and performs the "pick and place" task (20-50s). As shown in Fig.~\ref{fig:latency_performance}, the message delay exhibits minor fluctuations but keeps below 3ms, which is substantially lower than the 33ms interval of the 30Hz joint state data rate. Moreover, the std fluctuates slightly around only 4ms, implying that our attack is robust and hard to detect. Notably, we find that the message drop rate remains \textbf{0\%}. This indicates that our attack can robustly intercept and modify data in real time without delaying the intended task, seamlessly deceiving the victim's visualization interface (as shown in the supplementary video).

\subsection{Fidelity of the Synthesized Telemetry (RQ3)}
To quantify the discrepancy between the synthesized telemetry and the physical ground truth, we run the task 100 times, both with and without the attack, and compare their joint states using: 1) \textit{Root Mean Square Error} (RMSE) on positions, where a low RMSE indicates that the simulated kinematics accurately mirror the intended physical execution; 2) \textit{Dynamic Time Warping} (DTW) distance on velocities to measure the structural similarity of the motion; and 3) the \textit{std} of torques to capture the spread of mechanical loading. For each joint, we pair every synthesized run with a genuine run and temporally align them, meanwhile using Real-Real pairs of two distinct genuine runs as a genuine-variability reference.

Across all metrics, the synthesized telemetry falls within Real-Real reference variability. In Fig.~\ref{fig:position_rmse}, the Real-Fake position RMSE tracks the reference on every joint, with both dominated by the same shoulder and wrist joints (J2, J6) and no joint showing a separated median. Fig.~\ref{fig:velocity_dtw} shows the same velocity trend: the two distributions overlap and share an identical per-joint profile, the Real-Fake medians offset only marginally, yet well inside the reference spread. Fig.~\ref{fig:effort_std} confirms the torque scale is reproduced: the per-run std of the synthesized runs nearly coincides with the real one on every joint, recovering both the gravity-loaded shoulder (J2) and the near-zero distal joints (J5--J7). Beyond that, the per-joint medians and interquartile ranges of the Real-Fake and reference discrepancies coincide, leaving the downstream verifier no single-joint or single-channel statistic on which to separate faked from genuine motion.

\subsection{Multimodal Telemetry Synthesis (RQ4)}
To investigate if our attack can extend to other modalities, \ie, camera frames, we further integrate a VLM-based image synthesizer. When the victim invokes the camera service, our hook intercepts the legitimate frame and replaces it with a fake one. Specifically, after the simulated robot finishes the task, we use three images as the context (Fig.~\ref{fig:multimodal}): the simulated initial state from Isaac Sim (a, e), the intercepted image of the initial state (b, f) from a third-person perspective camera and a wrist camera, and the simulated final state (c, g). We then instruct the VLM to preserve the background environment of the real image while rendering the simulated robot's final state into a photorealistic equivalent and returning to the victim. We can see that the synthesized images (d, h) accurately render the kinematic state while, more importantly, also "placing" the object in the target place.

\subsection{Ablation Study (RQ5)}
To assess the contributions of each technical module for telemetry synthesis (\S\ref{sec:synthesize}), we consider four ablation settings: \ding{182} the raw synthesized telemetry, \ding{183} addition of the micro-motion, \ding{184} addition of the inter-joint coupling, and \ding{185} the full synthesis pipeline. We compute the RMSE, DTW, and std of the position, velocity, and torque between the ablated telemetry and the ground truth. In addition, we measure the bypass rate of the four ablation settings against the four verifiers (\S\ref{sec:overall}).

\begin{table*}[t]
\centering
\caption{Ablation study results (\ding{182} -- raw vs. real, \ding{183} -- raw + latent micro motion vs. real, \ding{184} -- raw + latent micro motion + inter-joint coupling vs. real, \ding{185} -- full synthesis vs. real).}
\label{tab:pairwise_metrics_real}
\scriptsize
\setlength{\tabcolsep}{3.5pt}
\begin{tabular}{@{}ccccccccc@{}}
\toprule
Metric & Pair & J1 & J2 & J3 & J4 & J5 & J6 & J7 \\
\midrule
\multirow{4}{*}{\shortstack[c]{Position\\(RMSE)}}
 & \ding{182} & $0.0420\pm0.0552$ & $0.1248\pm0.1782$ & $0.0442\pm0.0428$ & $0.0110\pm0.0077$ & $0.0217\pm0.0189$ & $0.1118\pm0.1472$ & $0.0458\pm0.0478$ \\
 & \ding{183}  & $0.0417\pm0.0551$ & $0.1209\pm0.1615$ & $0.0409\pm0.0427$ & $0.0108\pm0.0077$ & $0.0215\pm0.0188$ & $0.1109\pm0.1470$ & $0.0455\pm0.0478$ \\
 & \ding{184} & $0.0413\pm0.0550$ & $0.1118\pm0.1571$ & $0.0367\pm0.0426$ & $0.0107\pm0.0076$ & $0.0214\pm0.0185$ & $0.1105\pm0.1468$ & $0.0454\pm0.0477$ \\
 & \ding{185}  & $0.0412\pm0.0548$ & $0.1117\pm0.1502$ & $0.0349\pm0.0424$ & $0.0106\pm0.0075$ & $0.0214\pm0.0181$ & $0.1096\pm0.1464$ & $0.0451\pm0.0476$ \\
\midrule
\multirow{4}{*}{\shortstack[c]{Velocity\\(DTW)}}
 & \ding{182} & $0.0020\pm0.0002$ & $0.0029\pm0.0001$ & $0.0022\pm0.0002$ & $0.0012\pm0.0002$ & $0.0015\pm0.0002$ & $0.0038\pm0.0013$ & $0.0025\pm0.0004$ \\
 & \ding{183}  & $0.0019\pm0.0002$ & $0.0026\pm0.0001$ & $0.0020\pm0.0001$ & $0.0012\pm0.0002$ & $0.0014\pm0.0002$ & $0.0036\pm0.0008$ & $0.0024\pm0.0003$ \\
 & \ding{184} & $0.0019\pm0.0002$ & $0.0022\pm0.0001$ & $0.0020\pm0.0001$ & $0.0009\pm0.0002$ & $0.0013\pm0.0001$ & $0.0033\pm0.0004$ & $0.0023\pm0.0003$ \\
 & \ding{185}  & $0.0019\pm0.0002$ & $0.0019\pm0.0001$ & $0.0018\pm0.0001$ & $0.0008\pm0.0001$ & $0.0013\pm0.0001$ & $0.0029\pm0.0003$ & $0.0022\pm0.0004$ \\
\midrule
\multirow{4}{*}{\shortstack[c]{Torque\\(std)}}
 & \ding{182} & $0.0442\pm0.0152$ & $0.3618\pm0.3224$ & $0.0201\pm0.0147$ & $0.0264\pm0.0207$ & $0.0160\pm0.0125$ & $0.0169\pm0.0115$ & $0.0458\pm0.0366$ \\
 & \ding{183}  & $0.0412\pm0.0147$ & $0.3284\pm0.3223$ & $0.0198\pm0.0145$ & $0.0261\pm0.0206$ & $0.0155\pm0.0120$ & $0.0166\pm0.0112$ & $0.0456\pm0.0365$ \\
 & \ding{184} & $0.0406\pm0.0142$ & $0.3009\pm0.3223$ & $0.0197\pm0.0144$ & $0.0264\pm0.0207$ & $0.0147\pm0.0120$ & $0.0162\pm0.0113$ & $0.0458\pm0.0362$ \\
 & \ding{185}  & $0.0403\pm0.0142$ & $0.2793\pm0.3220$ & $0.0193\pm0.0143$ & $0.0262\pm0.0206$ & $0.0148\pm0.0123$ & $0.0153\pm0.0106$ & $0.0454\pm0.0363$ \\
\bottomrule
\end{tabular}
\label{tab:ablation}
\end{table*}

\begin{table}[t]
\centering
\setlength{\abovecaptionskip}{0pt}
\caption{The bypass rate of the four ablation settings.}
\begin{tabular}{ccccc}
\toprule
Pair & D1 & D2 & D3 & D4 \\
\midrule
\ding{182}    &  97\%  &  11\%  &  5\%  &  13\%  \\
\ding{183}    &  98\%  &  12\%  &  5\%  &  66\%  \\
\ding{184}    &  98\%  &  15\%  &  92\%  &  74\%  \\
\ding{185}    &  98\%  &  100\%  &  95\%  &  87\%  \\
\bottomrule
\end{tabular}
\label{tab:ablation_detector}
\end{table}

As shown in Table~\ref{tab:ablation}, the discrepancy between the synthesized and real telemetry decreases as each component is added, indicating that every module contributes to telemetry fidelity. Relative to the raw synthesized telemetry, the micro-motion restores the fine-grained fluctuations across all three channels. Inter-joint coupling provides a further reduction: the position RMSE of the gravity-loaded shoulder (J2) falls from 0.1209 to 0.1118. The full pipeline tightens the torque distribution through cross-channel consistency, lowering the J2 torque std from 0.3009 to 0.2793. These gains concentrate on the actively moving joints (J2, J6), whereas the near-static distal joints (J4, J5, J7) already match the real telemetry and change only within noise. Overall, the full synthesis achieves the closest match to real hardware, reducing J2 discrepancy by roughly 10\%, 34\%, and 23\% on position, velocity, and torque, respectively. Notably, as shown in Table~\ref{tab:ablation_detector}, only the full synthesis evades all four detectors. D1 is bypassed at $\sim$97\% throughout, as simulated telemetry already respects physical bounds. Inter-joint coupling drives D3 from 5\% to 92\%, while the micro-motion lifts D4 from 13\% to 66\%. D2 is evaded only by the full synthesis, whose trajectory derivation makes velocity exactly consistent with position. These results further confirm that each module is effective and necessary, together producing telemetry that no detector in the ensemble can distinguish from genuine hardware.

\section{Discussions \& Countermeasures}

\noindent\textbf{Live-Video Spoofing.}
Real-world robots often carry multiple cameras covering distinct viewpoints. In principle, fooling the verifier would require synthesizing coherent fake streams for all of them, but neither side can operate at that fidelity. Running the verifier on every live feed is prohibitively expensive at scale (\eg, a warehouse), and forging those feeds in real time is just as impractical for the attacker. We therefore model a weaker but realistic threat: forging only the final-state image, much like a food-delivery confirmation, where a single photo is enough to prove the delivery.

\noindent\textbf{Countermeasure.}
The root cause is that SROS~2 authenticates telemetry only \emph{after} \texttt{rcl\_publish}, downstream of the pre-publication boundary our hook occupies. A direct fix is to move the root of trust to the physical source: a trusted sensing element (\eg, a secure microcontroller or TEE) signs each payload with a monotonic counter at acquisition time, so a user-space hook can neither forge nor replay it.


\section{Conclusion}

We expose a fundamental trust boundary in ROS 2 between robot physical execution and the telemetry used to verify it. Entirely from user space, an adversary can preload a hook that intercepts messages before SROS 2 protects them, hijacking the robot while feeding verifiers synthesized telemetry authenticated as legitimate traffic. On a physical Franka arm, the attack succeeds in every trial while achieving real-time spoofing and evading representative detectors. This shows that securing telemetry in transit does not guarantee it reflects physical reality. Closing the gap requires anchoring trust at the physical source that signs each measurement at acquisition time.

\bibliographystyle{IEEEtran}

\end{document}